\documentclass[sigconf]{acmart} 
\usepackage{bookmark,xspace}
\usepackage{multirow}
\usepackage{paralist}
\usepackage{subfigure}
\usepackage{enumitem}
\usepackage{color}
\usepackage{algorithm}
\usepackage{algorithmic}

\AtBeginDocument{%
  \providecommand\BibTeX{{%
    \normalfont B\kern-0.5em{\scshape i\kern-0.25em b}\kern-0.8em\TeX}}}
\newcommand{\name}{IMF\xspace}

\copyrightyear{2023} 
\acmYear{2023} 
\setcopyright{rightsretained} 
\acmConference[WWW '23]{Proceedings of the ACM Web Conference 2023}{April 30--May 5, 2023}{Austin, TX, USA} 
\acmBooktitle{Proceedings of the ACM Web Conference 2023 (WWW '23), April 30--May 5, 2023, Austin, TX, USA} 
\acmDOI{10.1145/3543507.3583554} 
\acmISBN{978-1-4503-9416-1/23/04}

\usepackage{etoolbox}
\makeatletter
\patchcmd{\maketitle}{\@copyrightpermission}{
   \begin{minipage}{0.3\columnwidth}
     \href{https://creativecommons.org/licenses/by/4.0/}{\includegraphics[width=0.90\textwidth]{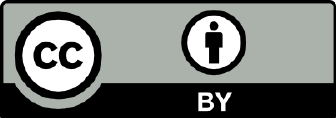}}
   \end{minipage}\hfill
   \begin{minipage}{0.7\columnwidth}
     \href{https://creativecommons.org/licenses/by/4.0/}{This work is licensed under a Creative Commons Attribution International 4.0 License.}
   \end{minipage}
  
   \vspace{5pt}
}{}{}
\makeatother

\begin{document}

\title{IMF: Interactive Multimodal Fusion Model for Link Prediction}

\author{Xinhang Li}
\affiliation{%
 \institution{Department of Computer Science and Technology, Tsinghua Univerisity}
 \city{Beijing}
 \country{China}
}
\email{xh-li20@mails.tsinghua.edu.cn}

\author{Xiangyu Zhao}
\authornotemark[1]
\affiliation{%
 \institution{School of Data Science, City Univerisity of Hong Kong}
 \country{Hong Kong}
}
\email{xianzhao@cityu.edu.hk}

\author{Jiaxing Xu}
\affiliation{%
 \institution{School of Computer Science and Engineering, Nanyang Technological Univerisity}
 \country{Singapore}
}
\email{jiaxing.xu@ntu.edu.sg}

\author{Yong Zhang}
\authornote{Xiangyu Zhao and Yong Zhang are corresponding authors.}
\affiliation{%
 \institution{Department of Computer Science and Technology, Tsinghua Univerisity}
 \city{Beijing}
 \country{China}
}
\email{zhangyong05@tsinghua.edu.cn}

\author{Chunxiao Xing}
\affiliation{%
 \institution{Department of Computer Science and Technology, Tsinghua Univerisity}
 \city{Beijing}
 \country{China}
}
\email{xingcx@tsinghua.edu.cn}

\renewcommand{\shortauthors}{Xinhang Li et al.}

\begin{abstract}
Link prediction aims to identify potential missing triples in knowledge graphs. 
To get better results, some recent studies have introduced multimodal information to link prediction.
However, these methods utilize multimodal information separately and neglect the complicated interaction between different modalities.
In this paper, we aim at better modeling the inter-modality information and thus introduce a novel \textbf{I}nteractive \textbf{M}ultimodal \textbf{F}usion (\textbf{IMF}) model to integrate knowledge from different modalities.
To this end, we propose a two-stage multimodal fusion framework to preserve modality-specific knowledge as well as take advantage of the complementarity between different modalities.
Instead of directly projecting different modalities into a unified space, our multimodal fusion module limits the representations of different modalities independent while leverages bilinear pooling for fusion and incorporates contrastive learning as additional constraints.
Furthermore, the decision fusion module delivers the learned weighted average over the predictions of all modalities to better incorporate the complementarity of different modalities.
Our approach has been demonstrated to be effective through empirical evaluations on several real-world datasets.
The implementation code is available online at https://github.com/HestiaSky/IMF-Pytorch.
\end{abstract}


\maketitle
\section{Introduction}

Knowledge Graph (KG) stores rich knowledge and is essential for many real-world applications, such as question answering~\cite{DBLP:conf/acl/YihCHG15,DBLP:conf/ijcai/ZhouYHZXZ18,DBLP:conf/wsdm/HuangZLL19}, urban computing~\cite{zhao2017modeling,zhao2022multi} and recommendation systems~\cite{DBLP:conf/cikm/WangZWZLXG18,DBLP:conf/kdd/Wang00LC19,chen2022knowledge}.
Typically, a KG consists of relational triples, which are represented as \textit{\textless head entity, relation, tail entity\textgreater}~\cite{DBLP:journals/pieee/Nickel0TG16}.
Nevertheless, KGs are inevitably incomplete due to the complexity, diversity and mutability of knowledge. 
To fix this gap, the problem of link prediction is studied so as to predict potential missing triples~\cite{DBLP:conf/nips/BordesUGWY13}.

\begin{figure}
	\centering
	\includegraphics[width=\linewidth]{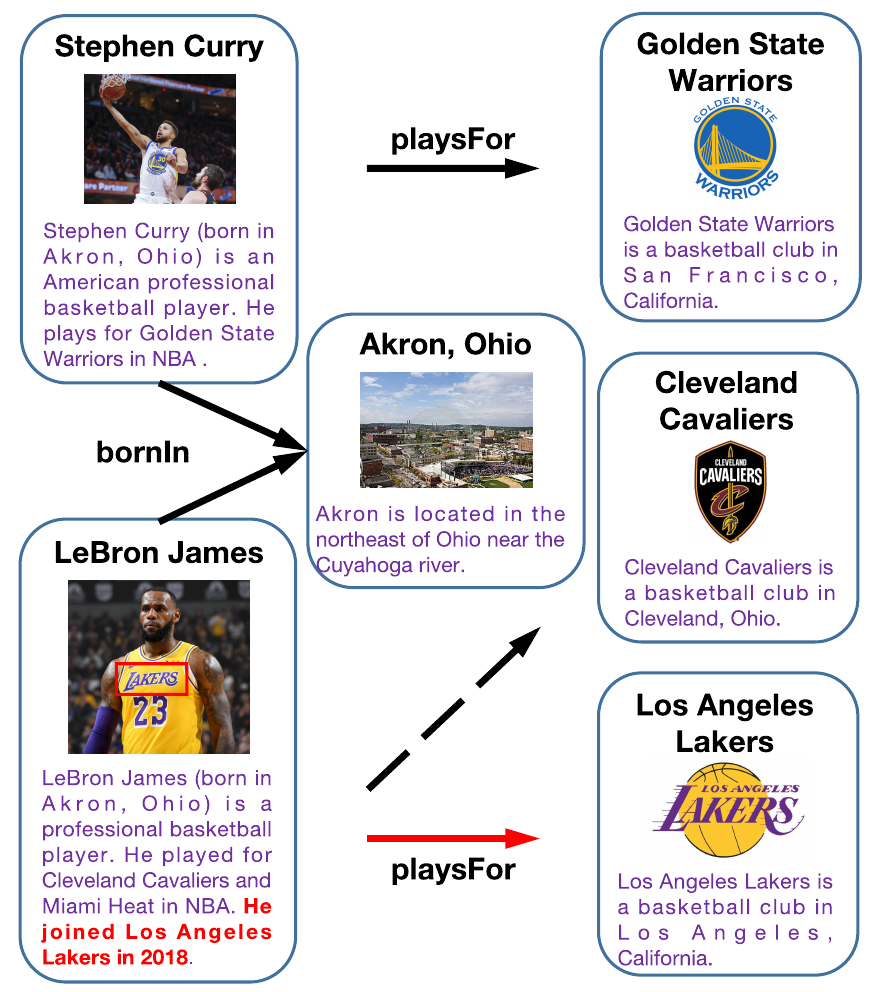}
	\caption{An example of link prediction which may be hard to predict without interaction of multimodal information.}
	\label{fig:example}
\end{figure}

Traditional link prediction models, including translation-based~\cite{DBLP:conf/nips/BordesUGWY13,DBLP:conf/aaai/WangZFC14} and neural network methods~\cite{DBLP:conf/naacl/NguyenNNP18,DBLP:conf/acl/NathaniCSK19}, suffered from the structural bias problem among triples.
Recently, some studies~\cite{DBLP:conf/ijcai/XieLLS17,DBLP:conf/starsem/SergiehBGR18,DBLP:conf/emnlp/PezeshkpourC018} addressed this problem by enriching the dataset and proposing new models to capture multimodal information for link prediction.
However, the performances of such studies were limited as they projected all modalities into a unified space with the same relation to capture the commonality, which might fail to preserve specific information in each modality.
As a result, they could not effectively model the complicated interactions between modalities to capture the complementarity.

To address the above issue, we incline to learn the knowledge comprehensively rather than separately, which is similar to how humans think.
Take the scenario in Figure~\ref{fig:example} as an example, such a model might also get the wrong prediction that \textit{LeBorn James} \texttt{playsFor} \textit{Golden States Warriors} based on the similarity with \textit{Stephen Curry} of the common \texttt{bornIn} relation to \textit{Akron, Ohio} in graph structure.
Meanwhile, it is difficult for visual features to express fine-grained semantics and the only conclusion is that \textit{LeBorn James} is a basketball player.
Also, it might also make the outdated prediction of \textit{Cleveland Cavaliers} due to `played' in the second sentence (more consistent with \texttt{playsFor} than `joined' in the third sentence) in the textual description.
Nevertheless, by integrating the knowledge, it is easy to get the correct answer \textit{Log Angeles Lakers} with the interaction between complementary information of structural, visual and textual highlighted in Figure~\ref{fig:example}.
Since the knowledge learned from different modalities is diverse and complex, it is very challenging to effectively integrate multimodal information.

In this paper, we propose a novel \textbf{I}nteractive \textbf{M}ultimodal \textbf{F}usion Model (\name) for multimodal link prediction over knowledge graphs. 
\name can learn the knowledge separately in each modality and jointly model the complicated interactions between different modalities with a two-stage fusion which is similar to the natural recognition process of human beings introduced above.
In the multimodal fusion stage, we employ a bilinear fusion mechanism to fully capture the complicated interactions between the multimodal features with contrastive learning.
For the basic link prediction model, we utilize the relation information as the context to rank the triples as predictions in each modality.
In the final decision fusion stage, we integrate predictions from different modalities and make use of the complementary information to make the final prediction.
The contributions of this paper are summarized as follows:
\begin{itemize}
    \item We propose a novel two-stage fusion model, \name, that is effective in integrating complementary information of different modalities for link prediction.
    \item We design an effective multimodal fusion module to capture bilinear interactions with contrastive learning for jointly modeling the commonality and complementarity.
    \item We demonstrate the effectiveness and generalization of \name with extensive experiments on four widely used datasets for multimodal link prediction.
\end{itemize}

\begin{figure*}[t]
    \centering
    \includegraphics[width=\linewidth]{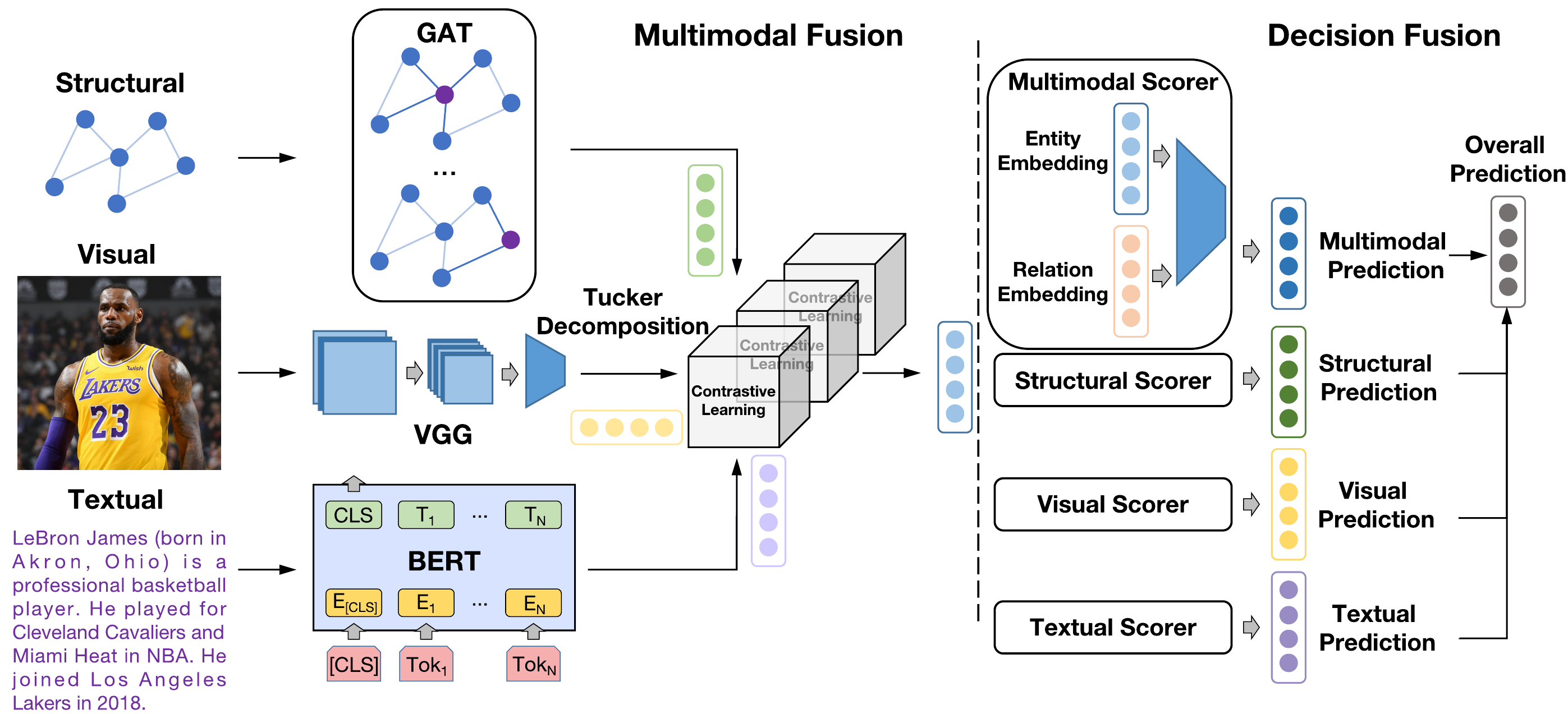}
    \caption{Overall architecture of \name. The left part represents different modality-specific encoders to extract latent features and the multimodal fusion module to integrate multimodal representations. The right part represents the contextual relational model decoders to get the similarity score and the decision fusion module to make the final prediction on all modalities.}
    \label{fig:model}
\end{figure*}

\section{Methodology}

Formally, a knowledge graph is defined as $\mathcal{G} = \langle \mathcal{E}, \mathcal{R}, \mathcal{T} \rangle$, where $\mathcal{E}$ and $\mathcal{R}$ indicate sets of entities and relations, respectively. 
$\mathcal{T} = \{(h, r, t) | h, t \in \mathcal{E}, r \in \mathcal{R}\}$ represents relational triples of the KG.
In multimodal KGs, each entity in KGs is represented by multiple features from different modalities.
Here, we define the set of modalities $\mathcal{K} = \{s, v, t, m\}$ where $s, v, t, m$ denote structural, visual, textual and multimodal modality, respectively.
Due to the complexity of real-world knowledge, it is almost impossible to take all the triples into account.
Therefore, given a well-formulated KG, the \emph{Link Prediction} task aims at predicting missing links between entities.
Specifically, link prediction models expect to learn a score function of relational triples to estimate the likelihood of a triple, which is always formulated as $\psi : \mathcal{E} \times \mathcal{R} \times \mathcal{E} \to \mathbb{R}$.

\subsection{Overall Architecture}

In order to fully exploit the complicated interaction between different modalities, we propose a two-stage fusion model instead of simply considering the multimodal information separately in a unified vector space.
As shown in Figure~\ref{fig:model}, \name consists of four key components:
\begin{itemize}[leftmargin=*]
	\item[1] The Modality-Specific Encoders are used for extracting structural, visual and textual features as the input of multimodal fusion stage.
	\item[2] The Multimodal Fusion Module, which is the first fusion stage, effectively models bilinear interactions between different modalities based on \textit{Tucker} decomposition and contrastive learning.
	\item[3] The Contextual Relational Model calculates the similarity of contextual entity representations to formulate triple scores as modality-specific predictions for decision fusion stage.
	\item[4]  The Decision Fusion Module, which is the second fusion stage, takes all the similarity scores from structural, visual, textual and multimodal models into account to make the final prediction.
\end{itemize}

\subsection{Modality-Specific Encoders}
In this subsection, we first introduce the pre-trained encoders used for different modalities.
These encoders are not fine-tuned during training and we treat them as fixed feature extractors to obtain the modality-specific entity representations.
Note that \name is a general framework and it is straightforward to replace them with other up-to-date encoders or add ones for new modalities into \name.

\subsubsection{Structural Encoder}

From the most basic view, the structural information of KG, we employ a Graph Attention Network (GAT)\footnote{https://github.com/Diego999/pyGAT}~\cite{DBLP:conf/iclr/VelickovicCCRLB18} with TransE loss.

Specifically, our GAT encoder takes L1 distance of neighbor aggregated representations as energy function of triples, which is $E(h, r, t) = ||\mathbf{h}+\mathbf{r}-\mathbf{t}||$.
In the training process, we minimize the following Hinge loss~\eqref{eq-gat-loss}:
\begin{equation}\label{eq-gat-loss}
    \begin{split}
        \mathcal{L}_{GAT} = & \sum_{(h,r,t) \in \mathcal{T}}\sum_{(h', r, t') \in \mathcal{T'}} \mathrm{max} \{0,  \\
        &\gamma + E(h,r,t) - E(h',r,t')\}
    \end{split}
\end{equation}
where $\gamma$ is margin hyper-parameter and $\mathcal{T'}$ denotes set of negative triples derived from $\mathcal{T}$. 
$\mathcal{T'}$ is created by randomly replacing head or tail entities of triples in $\mathcal{T}$, which is~\eqref{eq-gat-neg}:
\begin{equation}\label{eq-gat-neg}
    \mathcal{T'} = \{(h',r,t)|h' \in \mathcal{E} \backslash h\} \cup \{(h,r,t')|t' \in \mathcal{E} \backslash t\}
\end{equation}

\subsubsection{Visual Encoder} 
Visual features are greatly expressive while providing different views of knowledge from traditional KGs. 
To effectively extract visual features, we utilize VGG16\footnote{https://github.com/machrisaa/tensorflow-vgg} pre-trained on \textit{ImageNet}\footnote{https://image-net.org/} to get image embeddings of corresponding entities following~\cite{DBLP:conf/esws/LiuLGNOR19}.
Specifically, we take outputs of the last hidden layer before softmax operation as visual features, which are 4096-dimensional vectors.

\subsubsection{Textual Encoder} 
Entity descriptions contain much richer but more complex knowledge than pure KGs.
To fully extract the complex knowledge, we employ BERT~\cite{DBLP:conf/naacl/DevlinCLT19} as the textual encoder, which is very expressive to get description embeddings of corresponding entities.
The textual features are 768-dimensional vectors, i.e., pooled outputs of pre-trained BERT-Base model\footnote{https://github.com/huggingface/transformers}.

\subsection{Multimodal Fusion}
The multimodal fusion stage aims to effectively get multimodal representations, which fully capture the complex interactions between different modalities.
Many existing multimodal fusion methods have achieved promising results in many tasks like VQA (Visual Question Answering).
However, most of them aim at finding the commonality to get more precise representations by modality projecting~\cite{DBLP:conf/nips/FromeCSBDRM13,DBLP:conf/aaai/CollellZM17} or cross-modal attention~\cite{DBLP:conf/aaai/PerezSVDC18}.
These types of methods will suffer from the loss of unique information in different modalities and can not achieve sufficient interaction between modalities.
To this end, we propose to employ the bilinear models, which have a strong ability to realize full parameters interaction as the cornerstone to perform the fusion of multimodal information.
Specifically, we extend the \textit{Tucker} decomposition, which decomposes the tensor into a core tensor transformed by a matrix along with each mode to 4-mode factors as expressed in Equation~\eqref{eq-tucker}:
\begin{equation}\label{eq-tucker}
    \mathcal{P} = (((\mathcal{P}_c \times \mathbf{M}_s) \times \mathbf{M}_v) \times \mathbf{M}_t) \times \mathbf{M}_d
\end{equation}
where $\mathbf{M}_s \in \mathbb{R}^{d_s \times t_s}$, $\mathbf{M}_v \in \mathbb{R}^{d_v \times t_v}$, $\mathbf{M}_t \in \mathbb{R}^{d_t \times t_t}$,  $\mathbf{M}_d \in \mathbb{R}^{\mathcal{D} \times t_d}$ denotes transformation matrix and $\mathcal{P}_c \in \mathbb{R}^{t_s \times t_v \times t_t \times t_d}$ denotes a smaller core tensor.

In such a situation, entity embeddings are first projected into a low-dimensional space and then fused with the core tensor $\mathcal{P}_c$.
Following~\cite{DBLP:conf/iccv/Ben-younesCCT17}, we further reduce the computation complexity by decomposing the core tensor $\mathcal{P}_c$ to merge representations of all modalities into a unified space with element-wise product.
The detailed calculation process is expressed as Equation~\eqref{eq-fusion}:
\begin{equation}\label{eq-fusion}
    \mathbf{e}_m = \tilde{\mathbf{e}}_s^\mathsf{T} \mathbf{M}_d^s * \tilde{\mathbf{e}}_v^\mathsf{T} \mathbf{M}_d^v * \tilde{\mathbf{e}}_t^\mathsf{T} \mathbf{M}_d^t
\end{equation}
where $\tilde{\mathbf{e}}_k = \mathrm{ReLU}(\mathbf{e}_k\mathbf{M}_k) \in \mathbb{R}^{t_k}$ denotes latent representations and $\mathbf{e}_k \in \mathbb{R}^{d_k}$ is the original embedding representations and $\mathbf{M}_d^k \in \mathbb{R}^{t_k \times t_d}$ is decomposed transformation matrix for each modality $k \in \{s, v, t\}$.

\begin{figure}
    \centering
    \includegraphics[width=\linewidth]{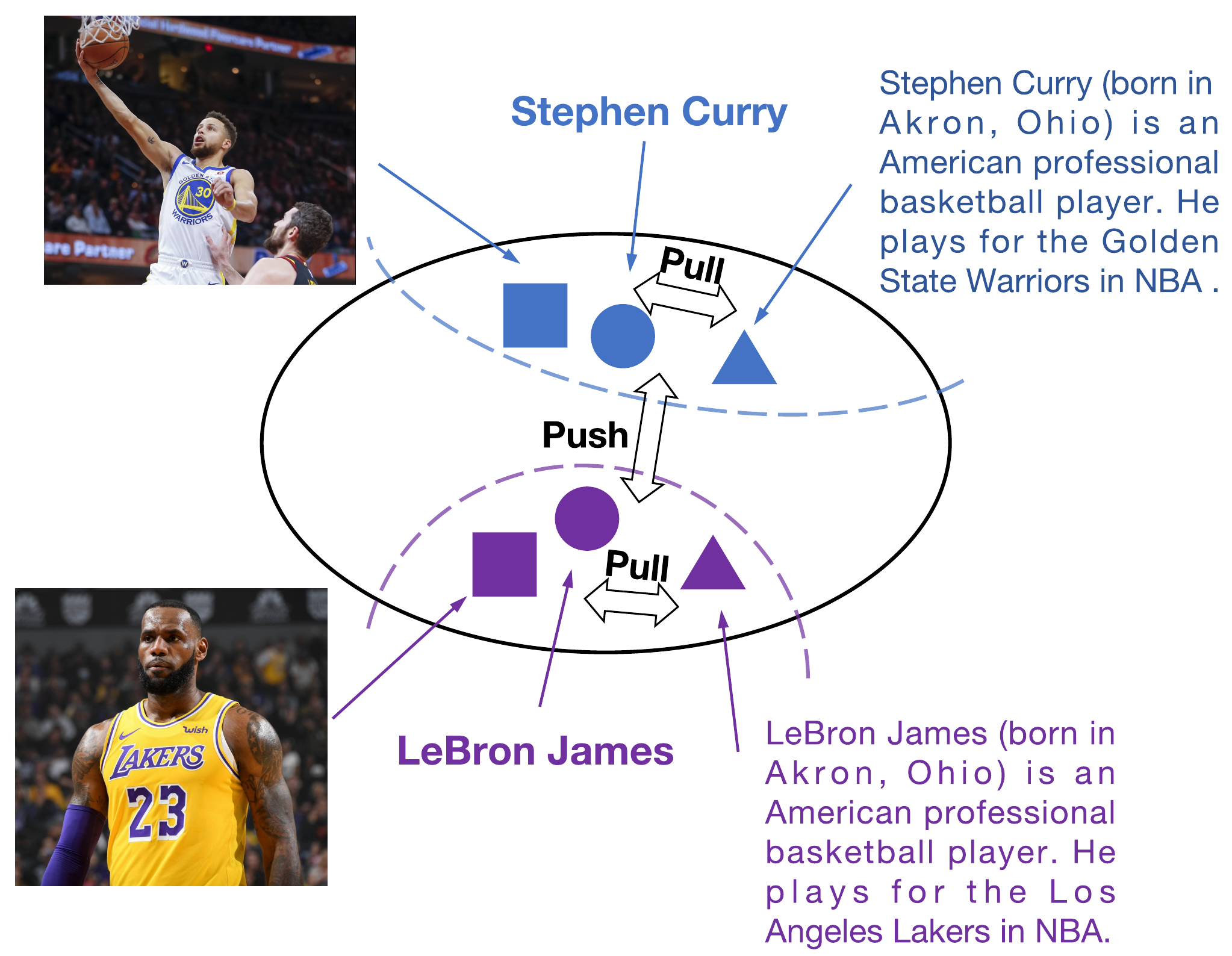}
    \caption{Example of multimodal contrastive learning. The distance between the representations of the same entity in different modalities is minimized, while the distance between the representations of different entities is maximized.}
    \label{fig:cl}
\end{figure}

However, the multimodal bilinear fusion has no bound limitation while the gradient produced by the final prediction result can only implicitly guide parameter learning.
To alleviate this problem, we add constraints to limit the correlation between different modality representations of the same entity to be stronger.
Therefore, we further leverage contrastive learning~\cite{DBLP:conf/icml/ChenK0H20,DBLP:conf/nips/LiSGJXH21,DBLP:conf/cvpr/Yuan0K0WMKF21} between different entities and modalities as an additional learning objective for regularization.
In the settings of contrastive learning, we take the pairs of representations of the same entity of different modalities as positive samples and the pairs of representations of different entities as negative samples.
As shown in Figure~\ref{fig:cl}, we aim at limiting the distance of negative samples to be larger than positive samples to enhance multimodal fusion, which is:
\begin{equation}
    d(f(x), f(x^+)) << d(f(x), f(x^-))
\end{equation}
where $d(\cdot, \cdot)$ denotes the distance measure and $f(\cdot)$ denotes the embedding function. The superscript $+, -$ represent the positive and negative samples, respectively.

Specifically, we randomly sample $N$ entities from the entity set as a minibatch and define contrastive learning loss upon it.
The positive pairs are naturally obtained with the same entities while the negative pairs are constructed by negative sharing~\cite{DBLP:conf/kdd/ChenSSH17} of all other entities.
We take the latent representations $\tilde{\mathbf{e}}_k = \mathrm{ReLU}(\mathbf{e}_k\mathbf{M}_k) \in \mathbb{R}^{t_k}$ and leverage cosine similarity $d(u, v) = - \mathbf{u}^\mathsf{T}\mathbf{v}/||\mathbf{u}||\mathbf{v}||$ as distance measure.
Then we have the following contrastive loss function for each entity $i$:
\begin{equation}\label{eq-cl}
    \mathcal{L}_{CLi} = \frac{1}{3N} \sum_{p,q \in \mathcal{M}} \sum_{j=1}^N  d(e_i^{p}, e_i^{q}) - d(e_i^{p}, e_j^{q}) + 2
\end{equation}
where $\mathcal{M} = \{(s, v), (s, t), (v, t)\}$ is set of modality pairs.

\subsection{Contextual Relational Model}
After obtaining representations of each modality and multimodal, we then design a contextual relational model, which takes relations in triples as contextual information for scoring, to get the predictions.
Note that this relational model can be easily replaced by any scoring function like TransE.

Due to the variety and complexity of relations in KGs, we argue that improving the degree of parameter interaction~\cite{DBLP:conf/aaai/VashishthSNAT20} is crucial for better modeling the relational triples.
The degree of parameter interaction means the calculation ratio of each parameter to all other parameters. 
For example, dot product could achieve $1/d$ degree while cross product could achieve $(d-1)/d$ degree.
Based on this assumption, we propose to use bilinear outer product between entity and relation embeddings to incorporate contextual information into entity representations.
Instead of taking relations as input as in previous studies, our contextual relational model utilizes relations to provide context in the transformation matrix of entity embeddings.
Then, entity embeddings are projected using the contextual transformation matrix to get \emph{contextual embeddings}, which are used for calculating similarity with all candidate entities.
The learning objective is to minimize the binary cross-entropy loss.
For each modality $k \in \mathcal{K}$, the computation details are shown as Equation~\eqref{eq-crm} to Equation~\eqref{eq-loss}:
\begin{gather}
    \hat{\mathbf{e}}_k = \mathbf{e}_k^\mathsf{T}\mathbf{W}_k^r  + \mathbf{b} = \mathbf{e}_k^\mathsf{T}\mathbf{W}_k\mathbf{r} + \mathbf{b}_k \label{eq-crm} \\
    \mathbf{y}_k = \sigma(\mathrm{cosine}(\mathbf{e}_k, \hat{\mathbf{e}}_k)) = \sigma (\frac{\mathbf{e}_k \cdot \hat{\mathbf{e}}_k}   
    {|\mathbf{e}_k| |\hat{\mathbf{e}}_k|}) \label{eq-sim} \\
    \mathcal{L}_k = -\frac{1}{N} \sum_{i=1}^N (t_i \cdot \mathrm{log}(y_{i,k})+(1-t_i) \cdot \mathrm{log}(1-y_{i,k})) \label{eq-loss}
\end{gather}
where $\mathbf{e}_k$ and $\hat{\mathbf{e}}_k$ are original and contextual entity embeddings respectively;
$\mathbf{W}_k^r = \mathbf{W}_k \mathbf{r}$ denotes contextual transformation matrix which is obtained by matrix multiplication of weight matrix $\mathbf{W}_k$ and relation vectors $\mathbf{r}$ while $\mathbf{b}_k$ is a bias vector;
$\sigma$ is sigmoid function and $\mathbf{y}_k = [y_{1,k},y_{2,k},...,y_{N,k}]$ is final prediction of modality $k$.

\subsection{Decision Fusion}
Existing multimodal approaches mainly focus on projecting different modality representations into a unified space and predicting with commonality between modalities, which will fail to preserve the modality-specific knowledge.
We alleviate this problem in the decision fusion stage by joint learning and combining predictions of different modalities to further leverage the complementarity.

Under the multimodal settings, we assign different contextual relational models for each modality and utilize their own results for training in different views.
Recall the contrastive learning loss in Equation~\eqref{eq-cl}, the overall training objective is to minimize the joint loss shown in Equation~\eqref{eq-mmloss}:
\begin{equation}\label{eq-mmloss}
    \mathcal{L}_{Joint} = \gamma_s \mathcal{L}_s + \gamma_v \mathcal{L}_v + \gamma_t \mathcal{L}_t + \gamma_m \mathcal{L}_{m} + \mathcal{L}_{CL}
\end{equation}
where $\mathcal{L}_k$ denotes binary cross entropy loss for modality $k$ as Equation~\eqref{eq-loss} and $\gamma_k$ is a learned weight parameter.

\begin{algorithm}[t]
\caption{Optimization Algorithm.}\label{alg:optim}
\begin{algorithmic}[1]
\STATE \textbf{Input:} Multimodal Knowledge Graph $\mathcal{G}$
\STATE \textbf{Output:} Trained Model $\mathcal{M}$
\STATE Pre-train structural encoder GAT on $\mathcal{G}$ with the loss in Equation(1)
\STATE Obtain pre-trained visual encoder VGG16 and textual encoder BERT-base
\STATE Initialize the entity embeddings $\mathbf{E}_s, \mathbf{E}_v, \mathbf{E}_t$ in $\mathcal{M}$ with the outputs of pre-trained encoders
\WHILE{not converge}
    \STATE Sample a batch of entities from $\mathcal{G}$
    \FOR{Entity $e$ in batch}
    \STATE Obtain the structural, visual, textual embeddings $\mathbf{e}_s, \mathbf{e}_v, \mathbf{e}_t$ of entity $e$
    \STATE Compute the multimodal fused embeddings $\mathbf{e}_m$ of entity $e$ with Equation (4)
    \STATE Compute the contrastive learning loss $\mathcal{L}_{CL}$ with Equation (6)
    \STATE Compute the loss $\mathcal{L}_s, \mathcal{L}_v, \mathcal{L}_t, \mathcal{L}_m$ with modality-specific scorers via Equation (7) - Equation (9)
    \STATE Compute the joint loss $\mathcal{L}_{Joint}$ with the above losses $\mathcal{L}_s, \mathcal{L}_v, \mathcal{L}_t, \mathcal{L}_m, \mathcal{L}_{CL}$ via Equation (10)
    \STATE Update model parameters of $\mathcal{M}$ by minimizing $\mathcal{L}_{Joint}$
    \ENDFOR
\ENDWHILE
\RETURN $\mathcal{M}$
\end{algorithmic}
\end{algorithm}

To better illustrate the whole training process of \name, we describe it via the pseudo-code of the optimization algorithm.
As shown in Algorithm~\ref{alg:optim}, we first obtain the pre-trained encoders of structural, visual and textual and utilize them for entity embeddings (line 3-5).
Since the pre-trained models are much larger and more complex than \name, they are not fine-tuned and their outputs are directly used as inputs of \name.
The multimodal embeddings are obtained by multimodal fusion while contrastive learning is applied to further enhance the fusion stage (line 9-11).
During training, each modality delivers its own prediction and loss via the modality-specific scorers (line 12), and then the joint prediction and loss are computed based on all modalities including multimodal ones (line 14).

For inference, we propose to jointly consider the predictions of each modality as well as multimodal ones.
Specifically, the overall predictions are shown in Equation~\eqref{eq-df}:
\begin{equation}\label{eq-df}
    \mathbf{y}_{Joint} = \frac{\gamma_s \mathbf{y}_s + \gamma_v \mathbf{y}_v + \gamma_t \mathbf{y}_t + \gamma_m \mathbf{y}_m} {\gamma_s + \gamma_v + \gamma_t + \gamma_m}
\end{equation}
where $\gamma_k$ denotes weight for modality $k$ as same as Equation~\eqref{eq-mmloss} while the values in $\mathbf{y}$ are in [0, 1].

\section{Experimental Setup}

\subsection{Datasets}
In this paper, we use four public datasets to evaluate our model.
All the datasets consist of three modalities: structural triples, entity images and entity descriptions.
DB15K, FB15K and YAGO15K datasets are obtained from MMKG\footnote{https://github.com/nle-ml}~\cite{DBLP:conf/esws/LiuLGNOR19}, which is a collection of multimodal knowledge graph.
Specifically, we utilize the relational triples as structural features, entity images as visual features and we extract the entity descriptions from Wikidata~\cite{DBLP:journals/cacm/VrandecicK14} as textual features.
FB15K-237\footnote{https://www.microsoft.com/en-us/download/details.aspx?id=52312}~\cite{Toutanova2015ObservedVL} is a subset of FB15K, the visual and textual features in FB15K can be directly reused.
Each dataset is split with 70\%, 10\% and 20\% for training, validation and test.
The detailed statistics are shown in Table~\ref{tbl:datasets}.

In the process of evaluation, we consider four metrics of valid entities to measure the model performance following previous studies: (1) mean rank (MR); (2) mean reciprocal rank (MRR); (3) hits ratio (Hits@1 and Hits@10).

\begin{table}[t]
  \centering
  \scalebox{1}{
    \begin{tabular}{c|ccccc}
      \toprule
      \textbf{Datasets} & \textbf{\#Ent.} & \textbf{\#Rel.} & \textbf{\#Train} & \textbf{\#Valid} & \textbf{\#Test} \\
      \midrule
        DB15K & 14,777 & 279 & 69,319 & 9,903 & 19,806\\
        FB15K & 14,951 & 1,345 & 414,549 & 59,221 & 118,443\\
        YAGO15K & 15,283 & 32 & 86,020 & 12,289 & 24,577\\
        FB15K-237 & 14,541 & 237 & 272,115 & 17,535 & 20,466\\
      \bottomrule
    \end{tabular}
  }
  \caption{Statistics of datasets.}\label{tbl:datasets}
\end{table}

\begin{table*}[ht]
	\centering
	\scalebox{1}{
		\begin{tabular}{ccccc|cccc|cccc}
			\toprule
			& \multicolumn{4}{c}{DB15K} & \multicolumn{4}{c}{FB15K} & \multicolumn{4}{c}{YAGO15K}  \\
			& MR & MRR & H@1 & H@10 & MR & MRR & H@1 & H@10 & MR & MRR & H@1 & H@10 \\
			\midrule
			TransE & 1128 & 0.256 & 13.7 & 46.9 & 108 & 0.495 & 43.7 & 77.4 & 971 & 0.161 & 5.1 & 38.4 \\
			ConvE & 729 & 0.312 & 21.9 & 50.7 & 64 & 0.745 & 67.0 & 87.3 & 714 & 0.267 & 16.8 & 42.6 \\
			TuckER & 693 & 0.341 & 24.3 & 53.8 & 40 & 0.795 & 74.1 & 89.2 & 689 & 0.281 & 18.3 & 45.7\\
			\midrule
			IKRL & 984 & 0.222 & 11.1 & 42.6 & 83 & 0.594 & 48.4 & 76.8 & 854 & 0.139 & 4.8 & 31.7 \\
			MKGC & 981 & 0.208 & 10.8 & 41.9 & 79 & 0.601 & 49.2 & 77.1 & 939 & 0.129 & 4.1 & 29.7 \\
			MKBE & 747 & 0.332 & 23.5 & 51.3 & 48 & 0.783 & 70.4 & 87.8 & 633 & 0.273 & 17.5 & 42.3\\
			\midrule
			IMF (w/o MF) & 687 & 0.319 & 21.8 & 51.2 & 62 & 0.752 & 69.2 & 86.6 & 764 & 0.213 & 11.4 & 35.3\\ 
			IMF (w/o DF) & 541 & 0.443 & 38.1 & 57.3 & 51 & 0.791 & 73.9 & 90.1 & 527 & 0.297 & 21.3 & 46.3 \\ 
			IMF (w/o CL) & 483 & 0.481 & 42.3 & 59.9 & 29 & 0.833 & 78.1 & 90.8 & 501 & 0.289 & 20.5 & 45.9\\
			IMF & \textbf{478*} & \textbf{0.485*} & \textbf{42.7*} & \textbf{60.4*} & \textbf{27*} & \textbf{0.837*} & \textbf{78.5*} & \textbf{91.4*} & \textbf{488*} & \textbf{0.345*} & \textbf{27.6*} & \textbf{49.0*}\\ 
			\bottomrule
		\end{tabular}
	}
	\caption{Evaluation results on multimodal DB15K, FB15K and YAGO15K datasets from MMKG. ``\textbf{{\large *}}'' indicates the statistically significant improvements (i.e., two-sided t-test with $p<0.05$) over the best baseline.}\label{tbl:MMKG}
\end{table*}

\subsection{Baselines}
To demonstrate the effectiveness of our model, we choose two types of methods for comparison, which are monomodal methods and multimodal methods.

For monomodal models, we take the baselines including:
\begin{itemize}[leftmargin=*]
    \item \textbf{TransE}~\cite{DBLP:conf/nips/BordesUGWY13} defines relations as transformations between entities and designs an \emph{energy function} of relational triples as scoring function.
    \item \textbf{ConvE}~\cite{DBLP:conf/aaai/DettmersMS018} converts 1D entity and relation embeddings into 2D embeddings and utilizes Convolutional Neural Network (CNN) to model the interactions between entities and relations.
    \item \textbf{ConvKB}~\cite{DBLP:conf/naacl/NguyenNNP18} employs CNN on the concatenated embeddings of relational triples to compute the triple scores.
    \item \textbf{CapsE}~\cite{DBLP:conf/cikm/NguyenNNP20} utilizes Capsule Network~\cite{DBLP:conf/nips/SabourFH17} to capture the complex interactions between entities and relations for prediction.
    \item \textbf{RotatE}~\cite{DBLP:conf/iclr/SunDNT19} introduces rotation operations between entities to represent relations in the complex space to infer symmetry, antisymmetry, inversion and composition relation patterns.
    \item \textbf{QuatE}~\cite{DBLP:conf/nips/0007TYL19} extends rotation of the knowledge graph embeddings in the complex space into the quaternion space to obtain more degree of freedom.
    \item \textbf{KBAT}~\cite{DBLP:conf/acl/NathaniCSK19} leverages Graph Attention Network (GAT)~\cite{DBLP:conf/iclr/VelickovicCCRLB18} as encoder to aggregate neighbors and employs ConvKB as decoder to compute triple scores.
    \item \textbf{TuckER}~\cite{DBLP:conf/emnlp/BalazevicAH19} applies \textit{Tucker} decomposition to capture the high-level interactions between entity and relation embeddings.
    \item \textbf{HAKE}~\cite{DBLP:conf/aaai/ZhangCZW20} projects entities into polar coordinate system to model hierachical structures for incorporating semantics.
\end{itemize}

For multimodal models, we take the baselines including:
\begin{itemize}[leftmargin=*]
    \item \textbf{IKRL}~\cite{DBLP:conf/ijcai/XieLLS17} utilizes the TransE energy function as scoring function on each pair of modalities for joint prediction.
    \item \textbf{MKGC}~\cite{DBLP:conf/starsem/SergiehBGR18} extends IKRL with combination of different modalities to explicitly deliver alignment between modalities.
    \item \textbf{MKBE}~\cite{DBLP:conf/emnlp/PezeshkpourC018} employs DistMult~\cite{DBLP:journals/corr/YangYHGD14a} as scoring function and designs Generative Adversarial Network (GAN)~\cite{DBLP:conf/nips/GoodfellowPMXWOCB14} to predict missing modalities.
\end{itemize}

For the ablation study, we design three variants of \name: \name (w/o MF) utilizes only structural information; \name (w/o DF) simply takes multimodal representations for training and inference without decision fusion; \name (w/o CL) removes the contrastive learning loss.

\subsection{Implementation Details}

The experiments are implemented on the server with an Intel Xeon E5-2640 CPU, a 188GB RAM and four NVIDIA GeForce RTX 2080Ti GPUs using PyTorch 1.6.0. 
The model parameters are initialized with Xavier initialization and are optimized using Adam~\cite{DBLP:journals/corr/KingmaB14} optimizer.
The evaluation is conducted under the \textbf{R}\textbf{\footnotesize{ONDOM}} settings~\cite{DBLP:conf/acl/SunVSTY20}, where the correct triples are placed randomly in test set and the negative sampling are correctly employed without test leakage.

For DB15K, FB15K and YAGO15K, we obtain the results by running all the baselines with their released codes.
For FB15K-237, we directly obtain the results of TransE, ConvE, ConvKB, CapsE, RotatE, KBAT and TuckER from the re-evaluation work~\cite{DBLP:conf/acl/SunVSTY20} and run the models of QuatE, HAKE, IKRL, MKGC and MKBE with their released codes.

Note that the methods with other enhancing techniques, such as data augmentation~\cite{DBLP:conf/uai/BamlerSM19,zheng2022cbr,zheng2022ddr,li2022gromov} or AutoML~\cite{DBLP:conf/icde/ZhangYDC20,zhaok2021autoemb,zhao2021autoloss,lin2022adafs,wang2022autofield} are orthogonal to our approach for comparison. 

\begin{figure*}[ht]
    \centering
    \includegraphics[width=0.85\linewidth]{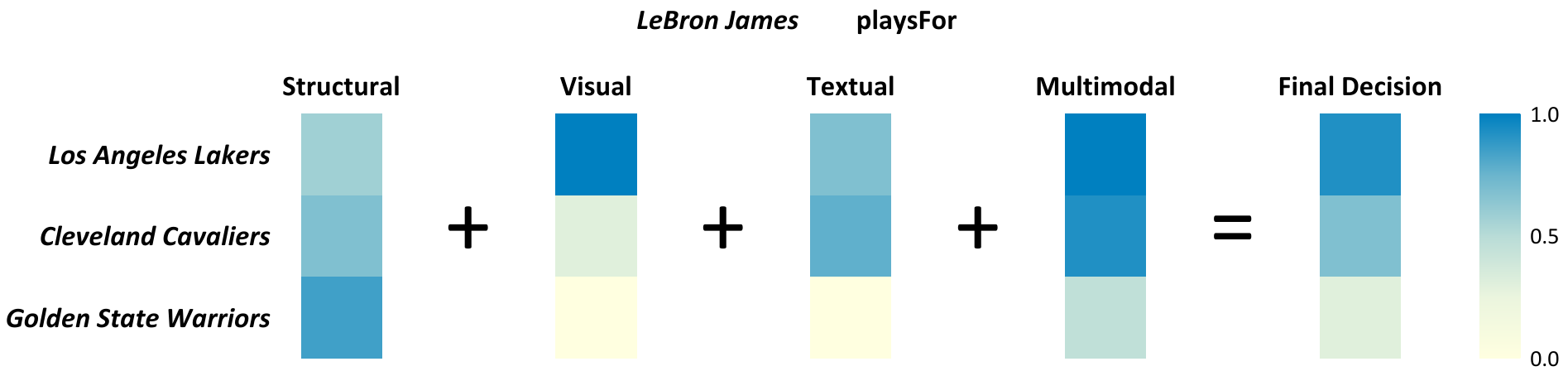}
    \caption{Visualization of prediction scores in decision fusion. }
    \label{fig:DF}
\end{figure*}

\section{Experimental Results}

\begin{table}
	\centering
	\scalebox{1}{
		\begin{tabular}{ccccc}
			\toprule
			& \multicolumn{4}{c}{FB15K-237}\\
			& MR & MRR & H@1 & H@10 \\
			\midrule
			TransE & 357 & 0.294 & - & 46.5\\
			ConvE & 244 & 0.325 & 23.7 & 50.1\\
			ConvKB & 309 & 0.243 & - & 42.1\\
			CapsE & 403 & 0.150 & - & 35.6\\
			RotatE & 177 & 0.338 & 24.1 & 53.3\\
			QuatE & 176 & 0.311 & 22.1 & 49.5\\
			KBAT & 223 & 0.232 & 13.6 & 42.8\\
			TuckER & 162 & 0.353 & 26.1 & 53.6\\
			HAKE & - & 0.346 & 25.0 & 54.2\\
			\midrule
			IKRL & 193 &  0.309 &  23.2 & 49.3\\
			MKGC & 187 &  0.297 &  22.9 & 49.4\\
			MKBE & 158 &  0.347 &  25.8 & 53.2\\
			\midrule
			IMF (w/o MF) & 188 & 0.324 & 23.4 & 51.8\\ 
			IMF (w/o DF) & 149 & 0.356 & 26.5 & 55.7\\ 
			IMF (w/o CL) & 138 & 0.371 & 27.8 & 57.1\\
			IMF & \textbf{134*} & \textbf{0.389*} & \textbf{28.7*} & \textbf{59.3*}\\ 
			\bottomrule
		\end{tabular}
	}
	\caption{Evaluation results on FB15K-237. ``*'' indicates the statistically significant improvements (i.e., two-sided t-test with $p<0.05$) over the best baseline.}\label{tbl:FB15K-237}
\end{table}

\subsection{Overall Performance}
As shown in Table~\ref{tbl:MMKG} and Table~\ref{tbl:FB15K-237}, we can observe that:

\begin{itemize}[leftmargin=*]
\item \name significantly outperforms all the baselines.
The performance gain is at most 42\% for MRR on DB15K while is also more than 20\% for all the evaluation metrics on average.

\item 
State-of-the-art monomodal methods employ a variety of complex models to improve the expressiveness and capture latent interactions.
However, the results illustrate that the performance is highly limited by the structural bias of the nature of knowledge graph itself.
Although these methods have already achieved promising results, \name can easily outperform them by a significant margin with a much simpler model structure, which amply demonstrates the effectiveness.

\item In comparison with multimodal methods that treat the features of different modalities separately, our \name jointly learning from different modalities with the two-stage fusion, which is beneficial in modeling the commonality and complementarity simultaneously.

\end{itemize}

Overall, our proposed \name can model more comprehensive interactions between different modalities with both commonality and complementarity thanks to the effective fusion of multimodal information and thus achieve significant improvement of link prediction on KGs.

\subsection{Ablation Study}

Table~\ref{tbl:MF} shows the evaluation results of leveraging different modality information on FB15K-237, where $S$ denotes structural information; $V$ denotes visual information of images and $T$ denotes textual information of descriptions.
We can see that by introducing visual or textual information, the performance is significantly improved.
The significant performance gain brought by multimodal fusion module not only demonstrates the effectiveness of our approach, but also indicates the potential of integrating multimodal information in KG.

\begin{table}
  \centering
  \scalebox{1}{
    \begin{tabular}{ccccc}
      \toprule
        & \multicolumn{4}{c}{FB15K-237}\\
        & MR & MRR & H@1 & H@10 \\
      \midrule
        S & 188 & 0.324 & 23.4 & 51.8\\ 
        S+V & 143 & 0.367 & 27.4 & 55.4\\ 
        S+T & 139 & 0.374 & 28.1 & 58.6\\ 
        S+V+T & \textbf{134} & \textbf{0.389} & \textbf{28.7} & \textbf{59.3}\\ 
      \bottomrule
    \end{tabular}
  }
  \caption{Evaluation results with different modality combinations on FB15K-237.}\label{tbl:MF}
\end{table}

To verify the effectiveness of decision fusion, we choose a case of \textless \textit{LeBron James}, \texttt{playsFor} \textgreater \ and visualize the prediction scores of each modality as Figure~\ref{fig:DF} shows.
Due to biases in each modality, the prediction of monomodal is inevitable error-prone.
The results in Table~\ref{tbl:MMKG} and Table~\ref{tbl:FB15K-237} also demonstrate the effectiveness of applying decision fusion to ensemble the specific latent features of each modality.

Besides, the performance comparison between \name (w/o CL) and \name in Table~\ref{tbl:MMKG} and Table~\ref{tbl:FB15K-237} illustrates the necessity of contrastive learning for more robust results, especially in the scenario with fewer training samples and relation types.

From the results shown above, we can see that each component in our propose \name has a significant contribution to the overall performance and it is beneficial to capture the commonality and complementarity between different modalities.

\subsection{Generalization}

\begin{figure}
	\centering
	\includegraphics[width=0.95\linewidth]{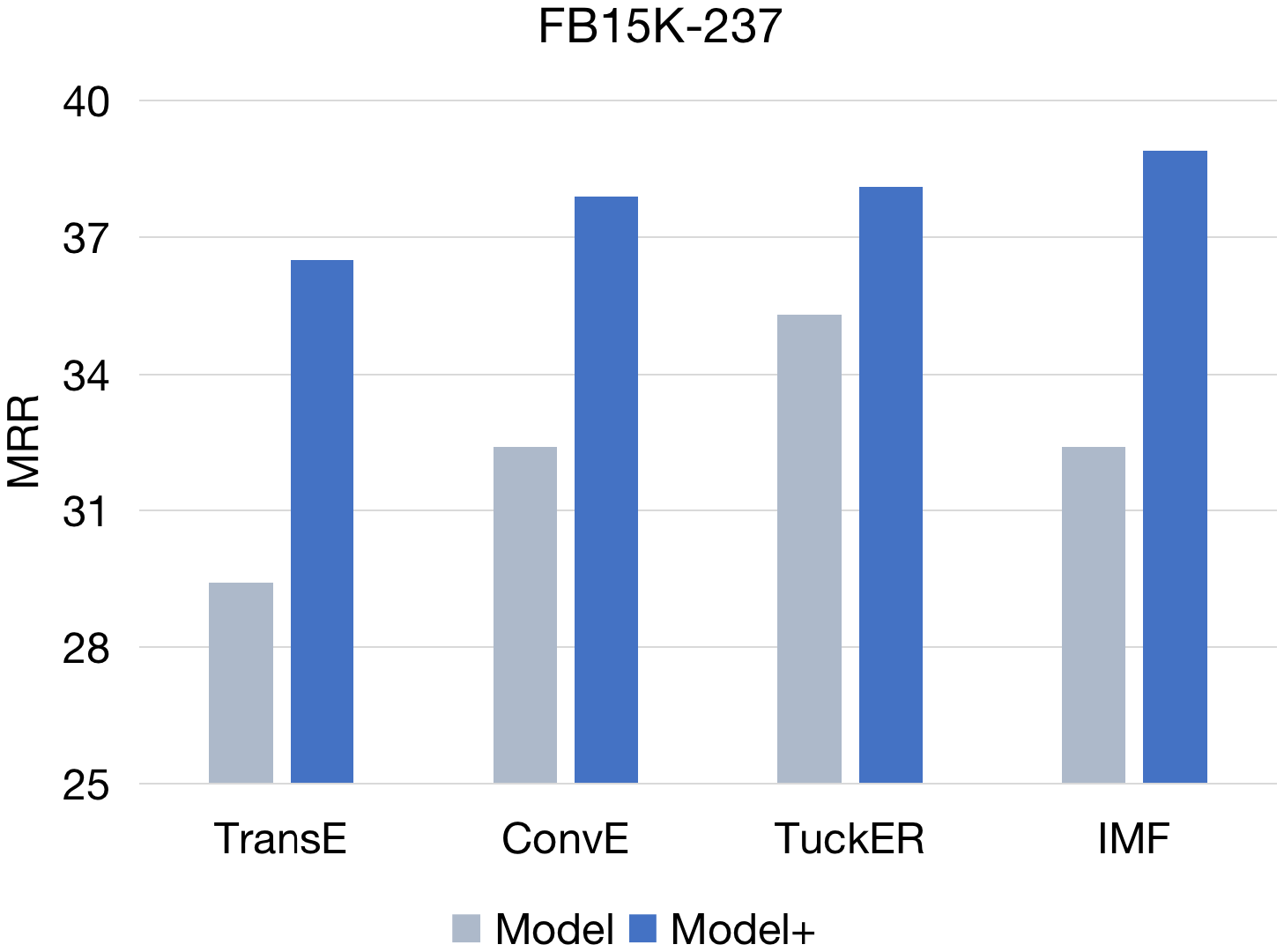}
	\caption{MRR (\%) improvement of different basic models on FB15K-237 with IMF.}
	\label{fig:TS}
\end{figure}

In order to evaluate the generalization of our proposed approach, we simply replace the scoring function (contextual relational model) with existing methods such as TransE, ConvE and TuckER.
The results in Figure~\ref{fig:TS} illustrate that our proposed framework of two-stage fusion is general enough to be applied to any link prediction model for further improvement.

\begin{figure*}
    \centering
    \includegraphics[width=0.62\linewidth]{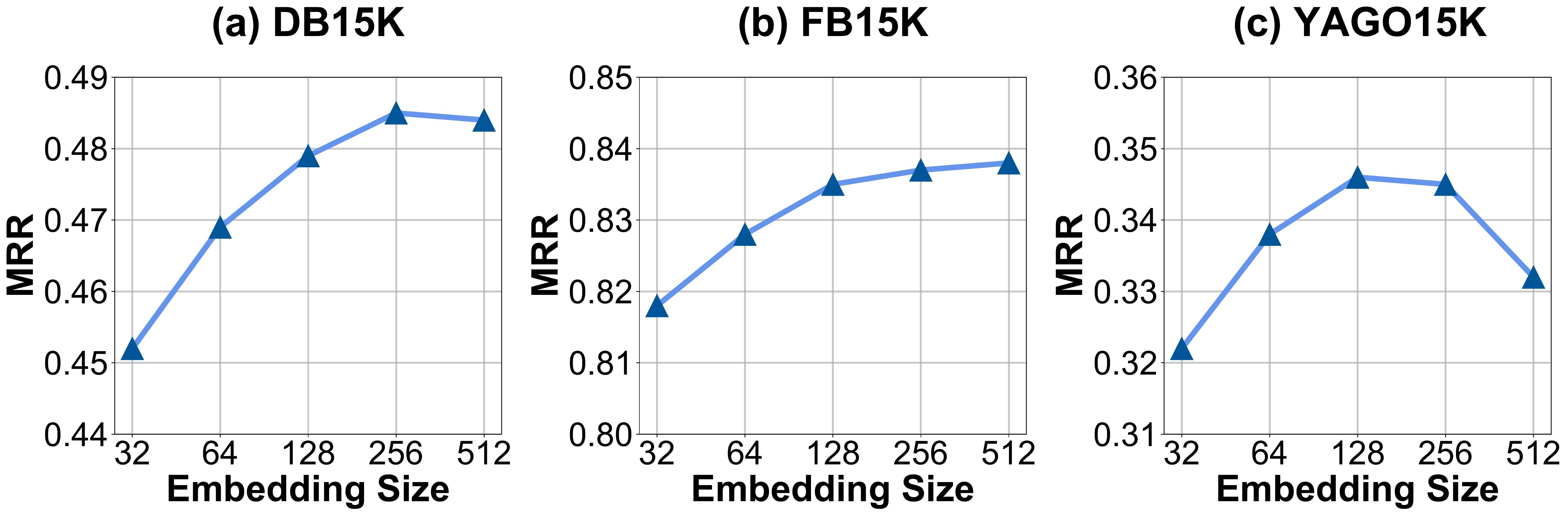}
    \caption{Performance influence of different embedding size.}
    \label{fig:param}
\end{figure*}

\begin{figure*}
	\centering
	\includegraphics[width=0.82\linewidth]{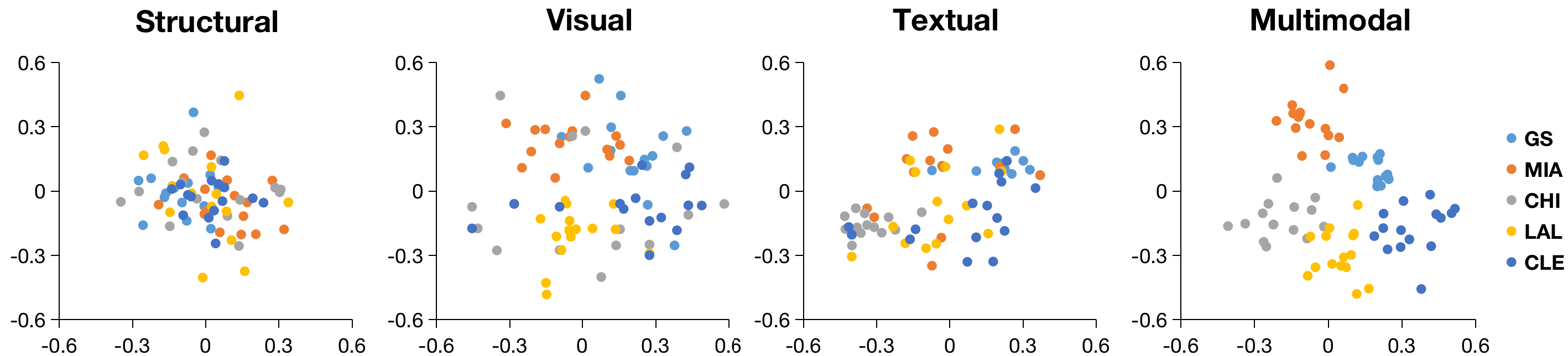}
	\caption{Visualization of low-dimensional representations for basketball players under the context \texttt{playsFor}. Each colored node denotes a basketball player and the different colors denote five basketball teams.}
	\label{fig:case}
\end{figure*}

\subsection{Parameter Analysis}

Figure~\ref{fig:param} shows the performance influence of embedding size for \name.
From the picture, we can see that the embedding size plays an important role in the model performance.
Meanwhile, it is worthy of note that a larger embedding size not always results in better performance due to the overfitting problem, especially in the datasets with fewer relation types like YAGO15K.
Considering the performance and the efficiency, the best choices of embedding size for these three datasets are 256, 256 and 128, respectively.

\subsection{Case Study}

In order to illustrate the effectiveness of our IMF model in a more intuitive way, we apply \textit{t-SNE} to reduce dimension and visualize the contextual entity representations of basketball players in five different basketball teams.
We can see in Figure~\ref{fig:case} that the representations of basketball players are messed up with monomodal information due to the biases.
However, with the help of interactive multimodal fusion, \name can effectively capture complicated interactions between different modalities.
\section{Related Work}

\subsection{Knowledge Embedding Methods}

Knowledge embedding methods have been widely used in graph representation learning tasks and have achieved great success on knowledge base completion (a.k.a link prediction).
Translation-based methods aim at finding the transformation relationships from source to target. 
TransE~\cite{DBLP:conf/nips/BordesUGWY13}, the most representative translation-based model, projects entities and relations into a unified vector space and minimizes the \emph{energy function} of triples. 
Following this route, many translation-based methods have emerged.
TransH~\cite{DBLP:conf/aaai/WangZFC14} formulates the translating process on relation-specific hyperplanes.
TransR~\cite{DBLP:conf/aaai/LinLSLZ15} projects entities and relations into separate spaces.

Recently, some neural network methods have shown promising results in this task. 
ConvE~\cite{DBLP:conf/aaai/DettmersMS018} and ConvKB~\cite{DBLP:conf/naacl/NguyenNNP18} utilize Convolutional Neural Network (CNN) to increase parameter interaction between entities and relations. 
KBAT~\cite{DBLP:conf/acl/NathaniCSK19} employ Graph Neural Networks (GNN) as the encoder to aggregate multi-hop neighborhood information.

However, all these methods above utilize only structural information, which is not sufficient for more complicated situations in real world.
By incorporating multimodal information in the training process, our approach is able to improve the representations with external knowledge.

\subsection{Multimodal Methods}

Leveraging multimodal information has yielded extraordinary results in many NLP tasks~\cite{DBLP:conf/iccv/Ben-younesCCT17}.
DeViSE~\cite{DBLP:conf/nips/FromeCSBDRM13} and Imagined~\cite{DBLP:conf/aaai/CollellZM17} propose to integrate multimodal information with modality projecting which learns a mapping from one modality to another.
FiLM~\cite{DBLP:conf/aaai/PerezSVDC18} extends cross-modal attention mechanism to extract textual-attentive features in visual models.
MuRel~\cite{DBLP:conf/cvpr/CadeneBCT19} utilizes pair-wise bilinear interaction between modalities and regions to fully capture the complementarity.
IKRL~\cite{DBLP:conf/ijcai/XieLLS17} is the first attempt at multimodal knowledge representation learning, which utilizes image data of the entities as extra information based on TransE.
MKGC~\cite{DBLP:conf/starsem/SergiehBGR18} combines textual and visual features extracted by domain-specific models as additional multimodal information compared to IKRL.
MKBE~\cite{DBLP:conf/emnlp/PezeshkpourC018} creates multimodal knowledge graphs by adding images, descriptions and attributes, and employs DistMult~\cite{DBLP:journals/corr/YangYHGD14a} as scoring function. 

Although these approaches did incorporate multimodal information to improve performance, they cannot take full advantage of it as they fail to effectively model interactions between modalities.

\section{Conclusion}

In this paper, we study the problem of link prediction over multimodal knowledge graphs.
Specifically, we aim at improving the interaction between different modalities.
To reach this goal, we propose the \name with a two-stage framework to enable effective fusion of multimodal information by (i) utilizing bilinear fusion to fully capture the complementarity between different modalities and contrastive learning to enhance the correlation between different modalities of the same entity to be stronger; and (ii) employing an ensembled loss function to jointly consider the predictions of multimodal representations.
Experimental results on several benchmarking datasets demonstrate the effectiveness of our proposed model.
Besides, we also conduct in-depth exploration to illustrate the generalization of our proposed method and the potential opportunity to apply it in real applications.

However, there are still some limitations of \name, which are left to future works.
For example, \name requires the integrity of all the modalities and an additional component to predict the missing modalities may be useful to tackle this limitation.
Besides, designing appropriate components to support more different kinds of modalities or propose a more lightweight fusion model to replace the bilinear model for better efficiency is also feasible.
\section*{ACKNOWLEDGEMENTS}
This research was partially supported by the National Key R\&D Program of China (No.2020AAA0109603), APRC - CityU New Research Initiatives (No.9610565, Start-up Grant for New Faculty of City University of Hong Kong), SIRG - CityU Strategic Interdisciplinary Research Grant (No.7020046, No.7020074), HKIDS Early Career Research Grant (No.9360163), Huawei Innovation Research Program and Ant Group (CCF-Ant Research Fund).

\bibliographystyle{ACM-Reference-Format}
\bibliography{custom}

\end{document}